# Towards autonomous system: flexible modular production system enhanced with large language model agents

Yuchen Xia
Institute of Industrial Automation
and Software Engineering
University of Stuttgart
Stuttgart, Germany
yuchen.xia@ias.uni-stuttgart.de

Manthan Shenoy
Institute of Industrial Automation
and Software Engineering
University of Stuttgart
Stuttgart, Germany
st175289@stud.uni-stuttgart.de

Nasser Jazdi
Institute of Industrial Automation
and Software Engineering
University of Stuttgart
Stuttgart, Germany
nasser.jazdi@ias.uni-stuttgart.de

Michael Weyrich
Institute of Industrial Automation
and Software Engineering
University of Stuttgart
Stuttgart, Germany
michael.weyrich@ias.uni-stuttgart.de

*Abstract* — In this paper, we present a novel framework that combines large language models (LLMs), digital twins and industrial automation system to enable intelligent planning and control of production processes. We retrofit the automation system for a modular production facility and create executable control interfaces of fine-granular functionalities and coarse-granular skills. Low-level functionalities are executed by automation components, and high-level skills are performed by automation modules. Subsequently, a digital twin system is developed, registering these interfaces and containing additional descriptive information about the production system. Based on the retrofitted automation system and the created digital twins, LLM-agents are designed to interpret descriptive information in the digital twins and control the physical system through service interfaces. These LLM-agents serve as intelligent agents on different levels within an automation system, enabling autonomous planning and control of flexible production. Given a task instruction as input, the LLM-agents orchestrate a sequence of atomic functionalities and skills to accomplish the task. We demonstrate how our implemented prototype can handle un-predefined tasks, plan a production process, and execute the operations. This research highlights the potential of integrating LLMs into industrial automation systems in the context of smart factory for more agile, flexible, and adaptive production processes, while it also underscores the critical insights and limitations for future work. Demos at: https://github.com/YuchenXia/GPT4IndustrialAutomation

## I. INTRODUCTION

Flexible production has emerged as a significant aspect of modern manufacturing environments in response to changing market demands and product customization requirements. Manufacturers need to adapt quickly to market changes and to stay competitive. This leads the manufacturer to consider diversifying their products and providing customized manufacturing services, which requires an agile production system and efficient management of the complexity of the production.

However, there are several technical challenges for deployment of agile and flexible production in reality: First of all, flexible production requires **seamless integration** of diverse technologies solution, e.g., robotics, automation, planning algorithms etc. Secondly, the production equipment and manufacturing processes need to be **reconfigurable** [1][2], which requires modular processes and systems as well as reconfigurable machines. Furthermore, automated flexible production also requires **quick changeover** [3] after decision-making to adapt the production against the changing requirements. Eventually, a highly **knowledgeable workforce** in every complicated technology with high availability to manage and supervise the complex system is too luxurious to be true. Traditional production systems frequently face difficulties in fulfilling these requirements due to their inflexible [1], dedicated workflows and restricted adaptability, as well as the absence of domain-specific knowledge in reconfiguring the production facility.

To tackle these challenges and requirements, we propose a novel solution: a large language model (LLM) enhanced automated modular production system for flexible manufacturing.

Our messages and contributions from this paper are summarized as follows:

(1) We demonstrate with a representative use case explaining why and how large language models can be used to achieve a higher level of intelligence and adaptability of industrial automation systems by planning and controlling the production, especially in the context of flexible production scenarios.

(2) We structure the system design according to the automation pyramid, illustrating a feasible technical approach to integrate LLMs into automation system.

(3) We prefer the more scalable in-context-learning approach over the fine-tuning approach, and the task-specific knowledge is injected into a LLM in prompt. As prompt engineering is an emerging field with little standardization, we devise a structured prompt template for this use case, drawing on insights from existing research in Natural Language Processing.

## II. BACKGROUND

In this section, we start by discussing why and how modular production systems can meet the requirements seamless integration and reconfigurability for flexible production. Then we emphasize the importance of modular query and control interfaces to allow the LLM to access information about the physical production processes and to adapt the production to changing requirements. Last but not least, we provide a brief overview of LLMs and the fundamental reasons why they have the potential to handle domain-specific tasks in industrial automation.

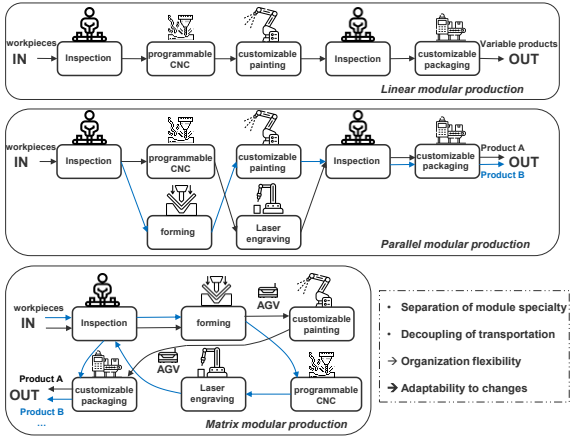

*Figure 1 Three structure types of modular production system*

### A. Modular Production Systems

Modular production systems are developed to address the challenges of flexibility, scalability, and adaptability in manufacturing. These systems consist of a series of modular modules, which can be easily reconfigured, replaced, or updated to accommodate varying production requirements. In the following introduction, we categorize them into three types: linear, parallel and matrix modular production.

#### 1) Linear modular production

The production process follows a step-by-step sequence, with each module performing its designated task before passing the workpiece to the next module. In discrete production like automotive assembly [4], the material can be processed differently with variable process module. In continuous production in the process industry, the process plants can be designed modular to decrease efforts and cost in system planning, integration, and configuration [5].

#### 2) Parallel modular production

The production system enables multiple modules and lines to operate concurrently on a variety of tasks. In comparison, parallel modular production supports simultaneous production operations, allowing a workpiece to be processed by multiple modules [4], which further increases flexibility. To effectively combine various production modules, additional transportation systems are necessary for seamless process automation.

#### 3) Matrix modular production

The matrix modular production [6] decouples the logistics tasks from production and changes the rigid line structure into matrix structure, which consist of modular production cells and automated transportation systems (often by applying **A**utomated **G**uided **V**ehicle). These systems comprise independent modules that can be reconfigured and combined to execute a wider range of production tasks. As various production modules with different specialties can be rearranged, added, or removed with minimal impact on the overall system, the matrix production has the potential to quickly adapt to diverse requirements, customer preferences, and market demands. Some literature also refers to this production type as Matrix Manufacturing Systems (MMS) [7].

Despite the structural superiority of matrix modular production for flexible reconfiguration, planning and process orchestration for customized production tasks still rely on the accumulated expertise within a company. Identifying a feasible solution to a problem can be time-consuming if any part of the required knowledge is unavailable or if there is a lack of effective communication among experts.

*Table 1 Comparison of different types of modular production against the changing customer demand examples*

| Case examples | Requirements | Linear MP | Parallel MP | Matrix MP |
|---|---|---|---|---|
| Customer wants the packaging material to be paper instead of plastics. | Variation of machine functionalities | + | + | + |
| Customer wants an engraved logo on the product instead of painted logo. | Variation of certain process steps | − | + | + |
| Customer wants a special HiL-quality test on the product in the middle of production process. | Variation of ochestration of processes | − | − | + |
| Customer returns the product due to a quality fault and demands reprocessing | Variation of problem-solving process | − | − | ○ |
| ＋: Requirements fulfilled without change of production system | | | | |
| −: Requirements hardly fulfilled due to the inflexible material flows | | | | |
| ○: Requirements can be fulfilled with experts intervene and effort | | | | |

Large Language Models (LLMs) possess the capability to interpret information, generate reasoning insights, and assist in decision-making processes. Trained on vast amounts of data, LLMs can understand and process complex information across various domains. By harnessing the interpretation and reasoning abilities of LLMs, the planning and process orchestration can be streamlined. This can lead to faster problem-solving and better adaptation to customer demands.

### B. Digital Twins

Despite the vast knowledge and reasoning capabilities of large language models (LLMs), a critical question remains: How can the LLMs access real-world information and effectively address tasks in practical settings?

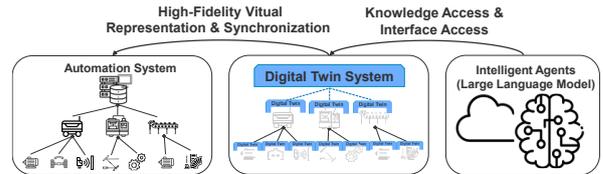

*Figure 2 The overall framework of the general concepts and their relationships*

The current state-of-the-art automation systems are not fully equipped to offer comprehensive descriptive information about production and unified accessible interfaces for querying and controlling physical processes. We developed a digital twin system to bridge the gap between LLMs and the physical world, as shown in Figure 2. Digital twins are synchronized virtual representations of physical assets or processes [8]. The digital twin system contains descriptive information about the production and exposes unified interfaces to LLM for manipulating the physical system. We lay special stress on the synchronization characteristics because it is fundamental to allow the reactive intelligent behavior of an autonomous system.

On one hand, automation systems are enhanced with digital twins and LLMs to unlock the potential of data- and AI-driven smart factories. On the other hand, LLMs interact with physical environments by having an embodiment in reality through the established infrastructure that combines automation systems and digital twins. This approach equips an artificial "brain" with mechatronic "hands" and "eyes" for more intelligent interaction.

### C. LLM in automated Production Systems

LLMs can be utilized to interpret complicated information, generate insights, and support decision-making processes in industrial automation systems.

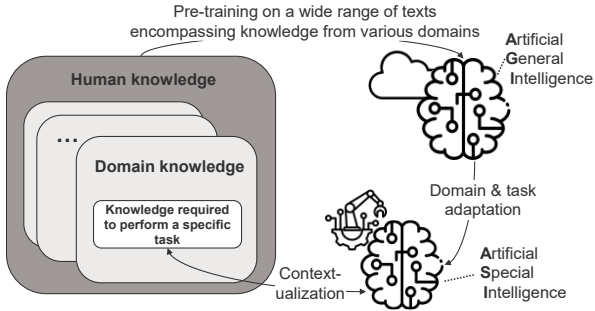

*Figure 4 Underlying general mechanism enabling LLMs to address domain-specific task.*

As illustrated in Figure 3, LLMs are deep learning models trained on vast amounts of text data, enabling them to generate human-like responses and handle complex language patterns across various NLP tasks. Recent advancements in NLP research have uncovered a promising finding: as the neuron size of LLMs increases, well-trained models gain the ability to interpret the meaning conveyed through language and demonstrate a capability of **approximating** human knowledge behind the language representation—a capability beyond the languages processing and not observed in smaller neural networks [9][10]. This development allows LLMs to perform general reasoning tasks effectively. Furthermore, as the training data for LLMs includes scientific papers, books, Q&A forums, and software code, LLMs are also informed with diverse domain-specific knowledge, which can be utilized for executing engineering related tasks.

By employing prompt engineering techniques [11], we develop multiple intelligent agents at both the MES level and the automation module level within the automation pyramid. These agents are specifically designed to manage production tasks within their respective scopes.

### III. METHODS

In this section, we explain how we connect the LLM to the digital twin infrastructure with prompt engineering, allowing intelligent agents to manage and control the production operations to solve an unforeseen problem.

### A. Integrate the information and expose the service interfaces of the digital twin

First and foremost, the large language model agent requires high-fidelity information to accurately comprehend the production system. Thus, a data infrastructure that houses comprehensive information about the production system is fundamental. We model the production system in a digital twin system in a modular and cascaded manner. These modular digital twins contain detailed information about their represented assets.

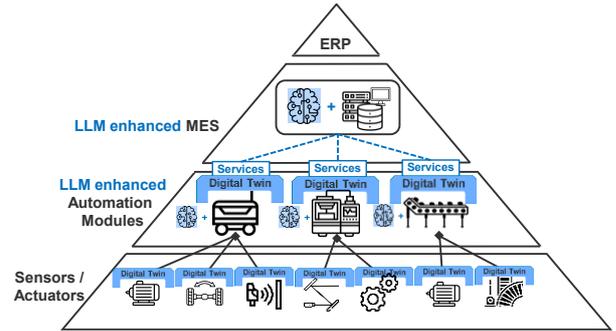

*Figure 3 Integration of LLM-agents and digital twins in automation systems for enhanced intelligence*

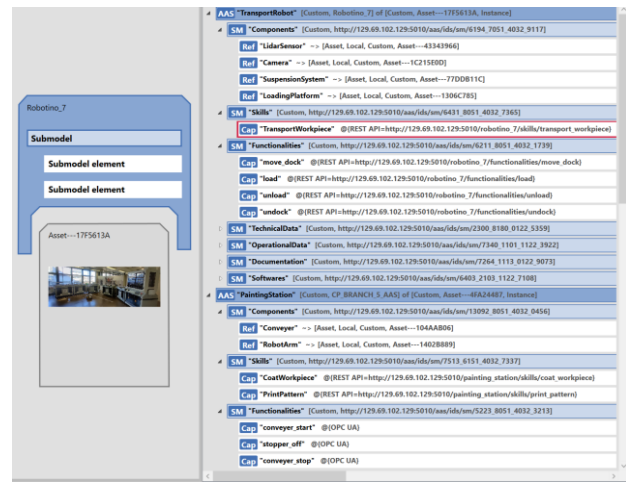

*Figure 5 Digital twin representation of the production system modeled with asset administration shell.*

The descriptive information in the digital twin system is modeled in the form of Asset Administration Shells (AAS) [1] and managed with an AAS-middleware [2]. Within the AAS, query and command services are referenced as URLs, which are semantically annotated with interface description in the skill sub-model. These interfaces enable the querying of asset states and control over automation system functionalities through RESTful service calls.

As shown in Figure 5, the digital twin of an automation module *"Transport Robot Robotino_7"* contains the cascaded information about its *"skills"*, the references to its *"components"*, callable *"functionalities"* interface, and other comprehensive information in sub-models *"technical data"* *"operational data" "documentation"* and *"software"* for further information.

Based on the descriptive knowledge about the assets and the callable interfaces provided by the digital twin system, it is possible to build two types of intelligent agents: A **manager**

---

[1] The Asset Administration Shell comprises extensive information pertaining to an asset, and it organizes this data into sub-models based on various aspects.

[2] We used Basyx AAS-middleware.

**agent** that works **on the top of** the automation modules, orchestrating diverse **skills** of the automation modules to plan the production; and several **operator agents** work **within** a particular automation module, orchestrating diverse **functionalities** to execute a given skill, as shown in Figure 5 and 6. Designing more than one agent is necessary to break down the challenging task into several manageable sub-tasks.

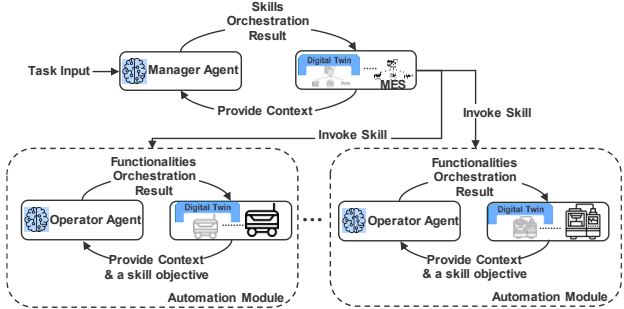

Figure 6 Interactions between LLM-agents and digital twins of automation modules and components

### B. Creating a LLM agent to adapt to a specific task with prompt engineering

We create these agents by contextualizing a GPT-model with prompts. The structure of the designed prompt and the interactions between digital twin and GPT-agent through the prompt are shown in Figure 7. More detailed examples are shown in the next *implementation section*.

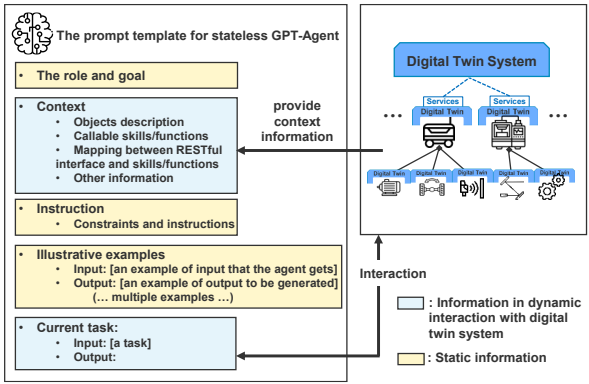

Figure 7 structure and content elements of the prompt and the interaction between the digital twin and GPT-agent through the prompt

The digital twin and the GPT-agents are connected via a prompt, which is sent to a LLM to initiate a response from the model. The prompt also serves as a trigger for the LLM to generate contextually appropriate text based on the given input. As shown in Figure 7, our designed prompts are composed of five distinct sections that target specific core working mechanisms of the GPT models. In the following texts, we explain them along with the design reasons. To increase the readability, we put further in-depth explanations related to NLP in footnotes.

#### 1) The role and the goal

This section of the prompt outlines the **role** and **objective** of an agent in natural language [3], providing the model with clarity regarding the particular task that is expected to carry out.

By defining the role and objective concisely, the text can effectively convey intentions and expectations with fewer tokens[4]. This offers two benefits: Firstly, it enables the model to better align with the general requirements, producing outputs that adhere to the desired role (e.g., operator, manager, advisor) and goal (e.g., performing operations, finding solutions, providing suggestions). Secondly, concise text concentrates meaning within fewer tokens, allowing the model to infer stronger related connections between text elements with higher attention weight [5] while preventing the dilution of the model's attention.

#### 2) Context

This section presents information derived from digital twins, aiming to supply descriptive information about the particular production system that the model needs for effective reasoning. As shown in table 2 and 3 in the *Implementation Section*, the knowledge should at least contain the objects description, the skills and functions of the objects and the mappings to the service interfaces of the executable operations. However, as the knowledge **representation** in digital twins' software system and texts in natural language are different, the information from the digital twin model shall be converted in text form in natural language, e.g., with fill-in-the-template mechanism and concatenation of text strings.

The converted information from digital twins offered in this context section serves two purposes: first, it enables the model to comprehend the production system's operations, incorporating additional information about the particular system. Secondly, as GPT has been trained across a wide range of subjects, it possesses extensive **general knowledge** that is implicitly stored within its model weights [12]. The descriptive information provided in the prompt guide GPT to "concentrate" [6] on the related knowledge embedded within the model when generating text. In this sense, this process actualizes the combination of the **general knowledge** of GPT with the **special knowledge** specified in prompt to execute reasoning for specific task.

#### 3) Instructions

This section aims to guide the GPT-agent's behavior by specifying the desired output formats and establishing boundaries for the generated content. We also encourage the model to "think step-by-step"[7], a widely adopted strategy among researchers, to facilitate logically structured reasoning

---

[3] This portion of the prompt targets the zero-shot learning capability [16], which enable large language models (LLMs) to perform previously unencountered tasks without explicit examples, particularly when the model has undergone instruct-fine-tuning optimization [17].
[4] A token can be defined as a basic meaningful unit of the text in input and output.
[5] Attention mechanism is analyzed and visualized in [18].
[6] This concentration is based on the mechanism of auto-regression process in text generation [13], where predicted text is continuously generated based on previous seen tokens.
[7] Also referred to as "*chain-of-thought-prompting*" [19] which significantly improves accuracy in performing complex reasoning [20][21]. Models that trained on code generation typically exhibit superior performance in step-by-step reasoning [19].

and break down complex problems into a series of smaller intermediate tasks.

*4) Illustrative examples*

This section provides verified concrete illustrative instances that demonstrate the desired input-output pattern. This can be beneficial in several ways: first of all, it constrains the structure of the text to be generated. Furthermore, the model's performance can be improved even with a limited number of examples[8]. Last but not least, the examples can help further specify the context information and disambiguate the abstract information provided so far, and during our experimenting, we observed an increase in misunderstood or irrelevant response from the model without representative instance examples.

*5) Input and Output interaction pattern*

While the previous prompt sections focus on configuring the LLM to comprehend the task, this prompt section focuses on instructing the LLM to generate texts based on specific input. The input can be a user request to the *"manager agent"* to perform production task by orchestrating the skills, or it could be a skill demand to an *"operator agent"* to perform a skill by orchestrating the functionalities within an automation module.

We intentionally leave the prompt incomplete, ending with *"Output:"*. By this means, the GPT's fundamental mechanism of next-token prediction [12][13] is addressed, upon which the model has been trained and optimized. Essentially, the agent carries out its designated task by completing the entire prompt and continuing writing the texts after the cue-word *"Output:"*.

## IV. IMPLEMENTATION AND EXPERIMENTS

In this section, we first illustrate our methods with two examples of the prompt, then explain the system components and their interaction with diagram and show the implemented demonstrator of a matrix modular production facility in our laboratory.

Table 2 and 3 contain minimum prompts we specified to prompt the GPT-model *"text-davinci-003"* to solve the production planning and execution problem. Readers can reproduce the results of agents output by sending the prompt text to the GPT-model for text generation[9].

*Table 2 Prompt input example of the stateless manager-agent.*

| The Prompt for the manager agent: |
|---|
| **Role and goal:** You are a manager of a production system. Your goal is to design an efficient production process based on a given task. You should take into account the provided context, instructions, and examples. Following these, you generate an output of a production process. |
| **Context:** <br> (1) A production process consists of one or more process steps. <br> (2) There are two types of process steps, one type is transportation process step, another type is production process step. <br> (3) If the next production process is executed in a different production module, transportation process between two production processes is necessary. <br> (4) The transportation step can be executed with a transport robot. <br> (5) Transportation step is not considered as production step. <br> (6) A production process always begins with a skill of the storage module and ends with a skill of the storage module. <br> (7) This production system that you manage consists of several production modules. Each of these production modules has one or more skills to execute a production process step. <br> (8) Each process step can be executed with one skill of a module. <br> (9) The production process should only contain the necessary steps that are necessary to satisfy a task specified in the input. <br> The production modules are described as following: <br> (10) An inspection module. It has the following skills: (I1) check the raw material, (I2) check the faulty material, (I3) test the quality of the material. <br> (11) A storage module. It has the following skills: (S1) retrieve a workpiece, (S2) store a workpiece. <br> (12) A transport robot. It has the following skills: (T1) transport workpiece between different modules. (T2) leave the production area. <br> (13) A CNC machine module. It has the following capabilities: (M1) drilling, (M2) milling, (M3) polishing. <br> (14) A painting module. It has the following skills: (P1) coat the material, (P2) print a pattern on the surface with paint. <br> (15) A laser machine module. It has the following skills: (L1) print a pattern on the surface with laser. |
| **Instructions:** As a manager of this production system, please arrange a production process based on the input. Only use the skills that are given in the context section. Only use the skills that are necessary to carry out the task. Give an explanation with short reason in list form. You should think step-by-step. |
| **Examples:** <br> Input: <br> {produce a steel sheet with a hole} <br> Output: <br> {(S1) – (T1) – (I1) – (T1) – (M1) – (T1) – (I3) – (T1) – (S2)} <br> Explanation: <br> (S1) retrieve a workpiece from storage module. <br> (T1) transport the workpiece from storage module to inspection module. <br> (I1) check the raw material to make sure it's suitable for production. <br> (T1) transport the material to the CNC machine module. <br> (M1) drill a hole on the steel sheet. <br> (T1) transport the material back to the inspection module. <br> (I3) test the quality of the material to ensure the final product meets the desired specifications. <br> (CT1) transport the material back to storage module to store the final product. <br> (…) <br> Input: {produce a steel nameplate and there should be a painted logo.} <br> Output: <br> {(S1) – (T1) – (I1) – (T1) – (M2) – (T1) – (L1) – (T1) – (I3) – (T1) – (S2)} <br> Explanation: <br> (S1) retrieve a workpiece from storage module. <br> (T1) transport the workpiece from storage module to inspection module. <br> (I1) check the raw material to make sure it's suitable for production. <br> (T1) transport the material to the CNC machine module. <br> (M2) mill the steel sheet to create the nameplate. <br> (T1) transport the material to the laser machine module. <br> (L1) print the logo on the nameplate using laser. <br> (T1) transport the material back to the inspection module. <br> (I3) test the quality of the material to ensure the final product meets the desired specifications. <br> (T1) transport the material back to storage module to store the final product. |
| Input: {the customer returned a wood nameplate and said there should be a painted customer logo on the backside. The wood nameplate is now in the storage module.} <br> Output: |
| **The generated output by the agent:** |
| {(S1) – (T1) – (P2) – (T1) – (I3) – (T1) – (S2)} <br> Explanation: <br> (S1) retrieve the wood nameplate from storage module. <br> (T1) transport the workpiece from storage module to painting module. <br> (P2) print the customer logo on the backside of the nameplate with paint. <br> (T1) transport the material back to the inspection module. <br> (I3) test the quality of the material to ensure the final product meets the desired specifications. <br> (T1) transport the material back to storage module to store the final product. |

*Table 3 Prompt input example of the stateless operator-agent for the transport robot*

| The Prompt for an operator agent (a transport robot): |
|---|
| **Role and goal:** You are an operator of a transport robot of a production system that performs skills to fulfill a transportation task. Your goal is to orchestrate the functionalities of this robot to perform a skill. You should take into account the provided context, instructions, and examples. Following these, you generate an output of a series of functionalities and provide the correct URLs to these functionalities. |
| **Context:** There are several production modules in this production system, and they are storage module, inspection module, painting module, CNC machine module and laser machine module. The transport robot only transports workpiece between two of the following modules: the inspection module, the painting module, the CNC machine module and the laser machine module. The transport robot can perform the following skill(s): (T1) Transport workpiece. A component named "functionality handler" is a logical component, and it is identified as "functionality_handler_001". It can control the actions of the transport robot. This component "functionality_handler_001" can execute the following functionalities of Robotino_7: <br> (1) Functionality "move_dock" will move the transport robot to a module and dock it to the module. This functionality can be called using the URL "http://129.69.102.129:5010/robotino_7/functionalities/move_dock". <br> (2) Functionality "load" will take a workpiece from a module and load it to the transport robot. This functionality can be called using the URL "http://129.69.102.129:5010/robotino_7/functionalities/load". <br> (3) Functionality "unload" will unload a workpiece from the transport robot and give it to a module. This functionality can be called using the URL "http://129.69.102.129:5010/robotino_7/functionalities/unload". <br> (4) Functionality "undock" will detach the transport robot from a module, if the transport robot is docked to a module. This functionality can be called using the URL "http://129.69.102.129:5010/robotino_7/functionalities/undock". |
| **Instructions:** As an operator of this transport robot, please arrange a series of functionalities to solve the task specified in the input. Only use the functionalities that are given. Only use the functionalities that are necessary to carry out the task. Specify the start module and target module in the output steps when necessary. Give an explanation with short reason in list form. You shall think step-by-step. Follow the text structure and syntax in the examples. |
| **Examples:** <br> Input: <br> (T1) Transport the workpiece from laser machine module to the CNC module. <br> Output: <br> To transport the workpiece from the laser machine module to the CNC module, the following steps shall be executed: |

---

| |
|---|
| (1) Move the transport robot to the laser machine module and dock it. Call the functionality "move_dock" using the URL "http://129.69.102.129:5010/robotino_7/functionalities/move_dock" to move the robot to the laser machine module and dock it.<br>(2) Load the workpiece from the laser machine module onto the transport robot. Call the functionality "load" using the URL "http://129.69.102.129:5010/robotino_7/functionalities/load" to load the workpiece onto the robot.<br>(3) Undock the transport robot from the laser machine module. Call the functionality "undock" using the URL "http://129.69.102.129:5010/robotino_7/functionalities/undock" to detach the robot from the laser machine module.<br>(4) Move the transport robot to the CNC module and dock it. Call the functionality "move_dock" using the URL "http://129.69.102.129:5010/robotino_7/functionalities/move_dock" to move the robot to the CNC module and dock it.<br>(5) Unload the workpiece from the transport robot onto the CNC module. Call the functionality "unload" using the URL "http://129.69.102.129:5010/robotino_7/functionalities/unload" to unload the workpiece from the robot onto the CNC module. |
| **Input:** {(T1) Transport the workpiece from the storage module to the painting module.}<br>**Output:** |
| **The generated output by the agent:** |
| To transport the workpiece from the storage module to the painting module, the following steps shall be executed:<br>(1) Move the transport robot to the storage module and dock it. Call the functionality "move_dock" using the URL "http://129.69.102.129:5010/robotino_7/functionalities/move_dock" to move the robot to the storage module and dock it.<br>(2) Load the workpiece from the storage module onto the transport robot. Call the functionality "load" using the URL "http://129.69.102.129:5010/robotino_7/functionalities/load" to load the workpiece onto the robot.<br>(3) Undock the transport robot from the storage module. Call the functionality "undock" using the URL "http://129.69.102.129:5010/robotino_7/functionalities/undock" to detach the robot from the storage module.<br>(4) Move the transport robot to the painting module and dock it. Call the functionality "move_dock" using the URL "http://129.69.102.129:5010/robotino_7/functionalities/move_dock" to move the robot to the painting module and dock it.<br>(5) Unload the workpiece from the transport robot onto the painting module. Call the functionality "unload" using the URL "http://129.69.102.129:5010/robotino_7/functionalities/unload" to unload the workpiece from the robot onto the painting module. |
| **The parsed result in JSON-file for control services invocation:**<br>**(by using regular expression to convert the texts into JSON in python)**[10] |
| [{"step": 1, "description": "Move the transport robot to the storage module and dock it", "action": "move_dock", "url": "http://129.69.102.129:5010/robotino_7/functionalities/move_dock"},{"step": 2, "description": "Load the workpiece from the storage module onto the transport robot", "action": "load", "url": "http://129.69.102.129:5010/robotino_7/functionalities/load"},{"step": 3, "description": "Undock the transport robot from the storage module", "action": "undock", "url": "http://129.69.102.129:5010/robotino_7/functionalities/undock"},{"step": 4, "description": "Move the transport robot to the painting module and dock it", "action": "move_dock", "url": "http://129.69.102.129:5010/robotino_7/functionalities/move_dock"},{"step": 5, "description": "Unload the workpiece from the transport robot onto the painting module", "action": "unload", "url": "http://129.69.102.129:5010/robotino_7/functionalities/unload"}] |

In the prompts, we allow updates to the information in the context section in accordance with the descriptive information and semantic annotations in the digital twin system. This is implemented in a python program with query-based template-filling and text string concatenation (cf. Figure 8). The program sends the prompt to GPT-model and parses the returned text into structured production steps with identifiers.

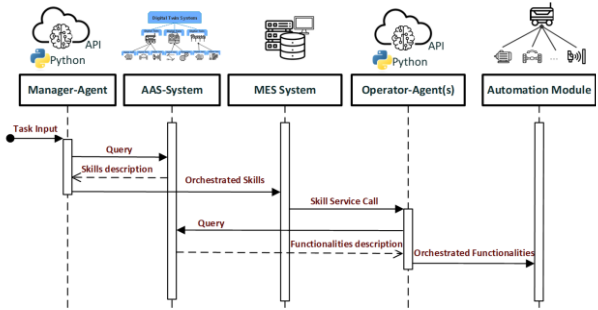

*Figure 9 Sequence diagram illustrating the interaction between different components of the prototype, from task input to operations execution.*

Several software components are implemented to realize our system design. The sequence diagram in Figure 8 shows the essential software components in our prototypical demonstrator and their interactions.

At the beginning of the process, a task input is handled by the manager agent. It queries a digital twin system (AAS-System) to retrieve the skills description of the automated production system and uses this data to update the context information in the specified prompt. The specified prompt is transmitted to the service API of a GPT-Model[11], and the GPT-model returns with the generated output texts. By parsing the generated texts, the manager agent obtains a sequence of skills to fulfill the task. The orchestrated skills are passed to the MES-System, based on which the MES-system invokes the skill service calls on one or more operator agent(s). On receiving the skill service calls, the operator agent retrieves the necessary information about the functionalities in an automation module from the digital twin system (AAS-System), and then orchestrates the functionalities to execute the requested skill. The orchestrated process is finally executed by the automation modules, as shown in Figure 9.

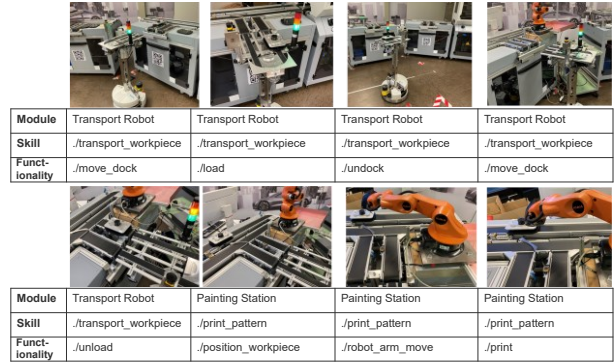

*Figure 8 Execution of production following the orchestrated processes directed by LLM-agents.*

## V. DISCUSSION

In the previous sections, we introduced a novel framework that integrates large language models (LLMs) with digital twin systems to enable intelligent management and control in industrial automation. Our method leverages prompt engineering to create LLM agents capable of adapting to specific tasks based on the information provided by digital twins. Through a case study involving a matrix modular production facility demonstrator, we showcased how our approach can address unforeseen problem tasks, autonomously orchestrate a production plan, and execute them to provide customizable production services.

Several key insights emerged from our study, and we explain them in three parts: positive insights, difficulties and lessons learned, as well as limitations and future works.

### A. Positive Insights

*1) Enhanced reasoning and decision-making capabilities for flexible/agile production*

By contextualizing the LLM agents using prompt engineering, we observed the human-like problem-solving capabilities in domain-specific tasks of production management and control.

Based on an instruct-finetuned model "*text-davinci-003*", we iteratively refined the proposed prompt template, which consists of five distinct sections. Using this template and the information provided by the digital twins, the defined LLM-agent effectively generates contextually appropriate responses to plan and control the production system autonomously. This reduces the need for human effort and can lead to increased productivity, reduced operational costs, and minimized delay time in production processes.

---

[10] Used regular expression in Python code:
\(((\d+)\) (.+)\. Call the functionality "(\w+)" using the URL "(.+)"

[11] We used the API of OpenAI's GPT-Model "text-davinci-003" for text generation in our prototypical implementation.

*2) Digital twin system bridges the information gaps for LLM-agent.*

As the traditional automation systems are not designed to host comprehensive information and knowledge about the production and the automation system itself, the development of a digital twin system (implemented with AAS) is necessary to provide the missing knowledge for the intelligent agents. The proposed architecture demonstrates the feasibility of establishing a bridge between LLMs and the physical production infrastructure with digital twin system. The modular and cascaded digital twins enable scalable communication, allowing the LLM-agents to access comprehensive information about the production system and interact with the physical world through service interfaces.

*3) Scalability, reuse of modular functionalities/skills to achieve adaptability*

Our approach demonstrates the necessities of developing scalable services interface of atomic functionalities/skills of automation components/modules. These interfaces are indispensable to enable intelligent LLM-agents to interact with the physical world. By dynamically orchestrating those atomic functionalities and skills, a higher level of adaptability of the production system can be achieved to meet the customized production demand.

*B. Difficulteis and Lessons Learned*

*1) Retrofitting the system to acquire modular interfaces.*
Retrofitting existing production facilities to accommodate modular interfaces can be a challenging process, necessitating time-consuming engineering efforts. While creating query interfaces is typically straightforward, developing control interfaces demands a higher level of caution due to the complexity involved. In comparison to coarse-granular skill, fine-granular functionalities exhibit more dependencies and require a closer examination of hardware-related aspects. Some required interfaces for our use case have application-level dependencies and cannot be executed independently without using the delivered vendor-specific software application. The modularity of the functionalities might not have been considered during the design phase when the system was developed. These dependencies across different components and on different system levels are preventing us from easily creating scalable interaction interfaces for our intelligent agents.

In order to solve this problem, we are looking into the code modularization on the device level to create the functionality interfaces from the bottom. By building from the bottom up, these "atomic" code-level functionalities can be orchestrated to create the modular skills of an automation module, allowing for greater reusability, flexibility and adaptability in system integration.

*2) Knowledge representation and conversion in prompts*
The conversion between knowledge representation in digital twins' software systems and natural language is a challenging task. Inaccurate or lossy conversion might result in inaccurate interpretation of the LLM agents for the production system and negatively affect their performance. Moreover, it is difficult to assess whether the LLM agents have accurately interpreted the context information provided in the prompt. How to determine the point at which refining the prompt would no longer yield significant performance improvement remains an open question. In our case study, we refined our prompts until the agents could generate effective results repeatedly (confer *section C.3.*).

When working on the iterative refinement of the prompts, we believe that the essential task is to "**translate the languages**" (e.g., translate the code and information models into text in natural language). In the translation, the knowledge conveyed by both representation forms shall pertain. Notably, we also use our own domain knowledge in automation and production engineering to design and additionally guide the LLM-agents. To be accurate in detail, LLM-agents don't understand the knowledge, but rather they **approximate the knowledge** conveyed by the representation.

*3) High-quality data and high-fidelity digital twins*
Supplying high-quality data and accurate knowledge representation in digital twin is essential to allow the LLM-agents to perform correct decision. However, creating high-fidelity digital twins can be labor-intensive and time-consuming, which also requires a comprehensive understanding of the system's architecture, behavior, and dependencies.

In our implementation, the digital twin system provides the asset information and exposes the updated interfaces description. The dynamic operational data are not replicated in digital twins to avoid data-inconsistency issues. The operational data should be provided by the MES system and the reverse-engineered RESTful-interfaces. In this sense, the digital twin system supplies descriptive information and annotations about the production system and components to help the LLMs to interpret how the production works and how to control it. However, we have not included the operational data into the prompt so far, because the GPT-agent has difficulty interpreting the dynamic numeric data and it would require extra memory for the model to cache the time series data. It is also due to the fact that we designed the LLM-agent to perform stateless interactions.

*C. Limitation and Future Work*

*1) Stateless interaction*
One notable limitation of our approach is the stateless interaction of the LLM agents because we only give all the context information at once through the API-call. The agent itself does not know how its output affects the production system. In order to keep track of the dynamic effects on environment and to perform more informed decision, the agent need to maintain a memory of the data about the production, which could require an extra software component (e.g., a database) or new mechanism to merge the LLM agents into the digital twin system.

*2) Non-deterministic results*
Another limitation of utilizing large language models (LLMs) is the inherent non-deterministic nature of the generated results. LLMs, such as GPT, are designed to predict the most probable next token in a sequence based on the input prompt and their extensive training data. Consequently, the outputs produced by LLMs can vary each time, even when the input remains the same. Although it is possible to set the model temperature to 0 to let the model generate more invariable output when given the same input, a decision-making process that only based on unexplained probability estimation would still be unreliable. As the task for customized production planning inherently involves non-deterministic input, additional mechanisms (extra constraints or guidance) are required to ensure a deterministic output and the predictability of the results.

*3) Data dilema for comprehensive testing and evaluation*

In contrast to general NLP tasks, where standard benchmarks and datasets are openly available, industrial automation tasks often involve heterogeneous data from specific components, complex hardware-level dependencies, a wide range of use cases, limited data collections, diversity of knowledge representation, and compliance with safety and reliability standards. Consequently, developing benchmarks for performance evaluation of LLM agents in these tasks necessitates future collaborative efforts. At this stage, we have not yet established an evaluation benchmark to comprehensively quantify the performance of LLMs in industrial automation tasks and can only provide a use-case-level proof-of-concept. For the given manager agent example in the *implementation section*, we evaluated 50 generated samples by the model "text-davinci-003": 96% of all the skill sequences are executable without error, 88% can solve the particular task and are able to produce the right product, however, only 6% of all the generated skills sequences use the minimal required steps to solve the task efficiently without unnecessary steps.

*4) The computational complexity of LLM models*

The most capable LLMs often come with considerable size and computational requirements, which can pose challenges for their local deployment in real-world industrial settings. Fine-tuning a smaller model with dedicated data for a specific domain could be a potential solution. However, we hypothesize that well-trained larger models with more neurons and trained on data from diverse knowledge domains possess a stronger general intelligence, which benefits the interpretation and reasoning in specialized domains and tasks. In other words, we assume that the larger models that perform better in general tasks may also yield better outcomes in domain-specific tasks. For simpler tasks, such as identifying semantic similarity in search queries or providing auto-completion recommendations [14][15], smaller embedding models are sufficient for functions with lower complexity.

## VI. CONCLUSION AND OUTLOOK

In conclusion, we have presented a novel framework that integrates large language models (LLMs) with digital twin systems for intelligent management and control in industrial automation. Our approach demonstrates the potential of LLM agents in making informed decisions based on the information provided by digital twins, leading to improved productivity, reduced operational costs, and minimized delay time in production planning processes.

Our study highlighted several positive insights, such as using LLMs to enhance the intelligence of the automation system, integrating of the LLMs into production system via digital twin systems, and the necessity of scalable and modular interfaces to increase the adaptability of production system. We also encountered challenges, including retrofitting existing systems, translating knowledge representation in prompts, and accommodating high-quality data and high-fidelity models with digital twins.

Last but not least, we identified limitations and future work directions, such as stateless interaction, non-deterministic results, data dilemma for benchmarking, and the computational requirements of LLMs. As the field of AI and industrial automation technology advances, we believe that integrating LLMs into industrial automation systems will lead to more efficient, flexible, and adaptive production systems. To further uncover and realize the potential of application of large language models in autonomous systems for industrial applications, it is crucial to engage in collaborative research efforts spanning across interdisciplinary fields.

## ACKNOWLEDGMENT

This work was supported by *Stiftung der Deutschen Wirtschaft (SDW)* and the Ministry of Science, Research and the Arts of the State of Baden-Wuerttemberg within the support of the projects of the *Exzellenzinitiative II*.

## REFERENCES

[1] T. Müller, B. Lindemann, T. Jung, N. Jazdi, and M. Weyrich, "Enhancing an Intelligent Digital Twin with a Self-organized Reconfiguration Management based on Adaptive Process Models," *Procedia CIRP*, vol. 104, pp. 786–791, Jan. 2021, doi: 10.1016/J.PROCIR.2021.11.132.

[2] T. Müller, N. Jazdi, J. P. Schmidt, and M. Weyrich, "Cyber-physical production systems: enhancement with a self-organized reconfiguration management," *Procedia CIRP*, vol. 99, pp. 549–554, Jan. 2021, doi: 10.1016/J.PROCIR.2021.03.075.

[3] M. Müller, T. Müller, B. Ashtari Talkhestani, P. Marks, N. Jazdi, and M. Weyrich, "Industrial autonomous systems: A survey on definitions, characteristics and abilities," *At-Automatisierungstechnik*, vol. 69, no. 1, pp. 3–13, Jan. 2021.

[4] P. Foith-Förster and I. Thomas Bauernhansl, "Changeable and reconfigurable assembly systems – A structure planning approach in automotive manufacturing," pp. 1173–1192, 2015, doi: 10.1007/978-3-658-08844-6_81.

[5] A. Markaj, A. Fay, N. Schoch, K. Stark, and M. Hoernicke, "Intention-based engineering for the early design phases and the automation of modular process plants," *IEEE International Conference on Emerging Technologies and Factory Automation, ETFA*, vol. 2022-September, 2022, doi: 10.1109/ETFA52439.2022.9921599.

[6] P. Greschke, "Matrix-Produktion als Konzept einer taktunabhängigen Fließfertigung," 2016.

[7] M. Trierweiler and T. Bauernhansl, "Reconfiguration of Production Equipment of Matrix Manufacturing Systems," pp. 20–27, 2021, doi: 10.1007/978-3-662-62962-8_3.

[8] D. Dittler, P. Lierhammer, D. Braun, T. Müller, N. Jazdi, and M. Weyrich, "An Agent-based Realisation for a continuous Model Adaption Approach in intelligent Digital Twins," 2022, Accessed: Mar. 21, 2023. [Online]. Available: https://arxiv.org/abs/2212.03681v1

[9] G. Jawahar, B. Sagot, and D. Seddah, "What Does BERT Learn about the Structure of Language?," *ACL 2019 - 57th Annual Meeting of the Association for Computational Linguistics, Proceedings of the Conference*, pp. 3651–3657, 2019, doi: 10.18653/V1/P19-1356.

[10] J. Wei *et al.*, "Emergent Abilities of Large Language Models," Jun. 2022, Accessed: Mar. 21, 2023. [Online]. Available: https://arxiv.org/abs/2206.07682v2

[11] T. Gao, A. Fisch, and D. Chen, "Making Pre-trained Language Models Better Few-shot Learners," *ACL-IJCNLP 2021 - 59th Ann. Meet. Assoc. Comput. Linguist. & 11th Int. Joint Conf. Nat. Lang. Process., Proceedings of the Conference*, pp. 3816–3830, 2021, doi: 10.18653/V1/2021.ACL-LONG.295.

[12] A. Radford and K. Narasimhan, "Improving Language Understanding by Generative Pre-Training," 2018.

[13] A. Vaswani *et al.*, "Attention Is All You Need," *Adv Neural Inf Process Syst*, vol. 2017-December, pp. 5999–6009, Jun. 2017, Accessed: Mar. 21, 2023. [Online]. Available: https://arxiv.org/abs/1706.03762v5

[14] M. Both, J. Muller, and C. Diedrich, "Reducing configuration efforts in energy management systems based on natural language processing methods and asset administration shells," *IEEE International Conference on Emerging Technologies and Factory Automation, ETFA*, vol. 2022-September, 2022, doi: 10.1109/ETFA52439.2022.9921479.

[15] Y. Xia, N. Jazdi, and M. Weyrich, "Automated generation of Asset Administration Shell: a transfer learning approach with neural language model and semantic fingerprints," *IEEE International Conference on Emerging Technologies and Factory Automation, ETFA*, vol. 2022-September, 2022, doi: 10.1109/ETFA52439.2022.9921637.

[16] V. Sanh *et al.*, "Multitask Prompted Training Enables Zero-Shot Task Generalization," Oct. 2021, Accessed: Mar. 21, 2023. [Online]. Available: https://arxiv.org/abs/2110.08207v3

[17] J. Wei *et al.*, "Finetuned Language Models Are Zero-Shot Learners," Sep. 2021, Accessed: Mar. 21, 2023. [Online]. Available: https://arxiv.org/abs/2109.01652v5

[18] J. Vig, Y. Belinkov, H. John, and A. Paulson, "Analyzing the Structure of Attention in a Transformer Language Model," pp. 63–76, Jun. 2019, doi: 10.18653/v1/w19-4808.

[19] J. Wei *et al.*, "Chain-of-Thought Prompting Elicits Reasoning in Large Language Models," Jan. 2022, Accessed: Mar. 21, 2023. [Online]. Available: https://arxiv.org/abs/2201.11903v6

[20] T. Kojima, S. Shane Gu, M. Reid Google Research, Y. Matsuo, and Y. Iwasawa, "Large Language Models are Zero-Shot Reasoners," May 2022, Accessed: Apr. 11, 2023. [Online]. Available: https://arxiv.org/abs/2205.11916v4

[21] B. Prystawski and N. D. Goodman, "Why think step-by-step? Reasoning emerges from the locality of experience," Apr. 2023, Accessed: Apr. 11, 2023. [Online]. Available: https://arxiv.org/abs/2304.03843v1

[22] T. B. Brown *et al.*, "Language Models are Few-Shot Learners," *Adv Neural Inf Process Syst*, vol. 2020-December, May 2020, Accessed: Mar. 21, 2023. [Online]. Available: https://arxiv.org/abs/2005.14165v4